\def\year{2022}\relax
\documentclass[letterpaper]{article} 
\usepackage{aaai22}  
\usepackage{times}  
\usepackage{helvet}  
\usepackage{courier}  
\usepackage[hyphens]{url} 
\usepackage{graphicx} 
\usepackage{natbib}  
\usepackage{caption} 
\DeclareCaptionStyle{ruled}{labelfont=normalfont,labelsep=colon,strut=off} 
\usepackage{algorithm}
\usepackage{algorithmic}
\usepackage{paralist}

\usepackage{amsmath,bm}
\usepackage{multirow}
\usepackage{amssymb}
\usepackage{booktabs}
\usepackage{nameref}
\usepackage{setspace}
\usepackage{hyperref}
\usepackage{fancyhdr}
\usepackage{newfloat}
\usepackage{listings}
\floatstyle{ruled}
\newfloat{listing}{tb}{lst}{}
\floatname{listing}{Listing}

\title{Search and Learning for Unsupervised Text Generation}
\author {
    Lili Mou 
}
\affiliations {
Department of Computing Science \& Alberta Machine Intelligence Institute (Amii)\\
University of Alberta, Canada CIFAR AI Chair\\
\url{doublepower.mou@gmail.com}
}
\begin{document}

\maketitle

\begin{abstract}
With the advances of deep learning techniques, text generation is attracting increasing interest in the artificial intelligence (AI) community, because of its wide applications and because it is an essential component of AI. Traditional text generation systems are trained in a supervised way, requiring massive labeled parallel corpora. In this paper, I will introduce our recent work on search and learning approaches to unsupervised text generation, where a heuristic objective function estimates the quality of a candidate sentence, and discrete search algorithms generate a sentence by maximizing the search objective. A machine learning model further learns from the search results to smooth out noise and improve efficiency. Our approach is important to the industry for building minimal viable products for a new task; it also has high social impacts for saving human annotation labor and for processing low-resource languages.

\end{abstract}

\newcommand{\newcite}[1]{\citeauthor{#1}~(\citeyear{#1})}

\renewcommand{\normalsize}{\fontsize{11pt}{0}\selectfont} 
\singlespacing

\thispagestyle{fancy}
\renewcommand{\headrulewidth}{0pt}
\rhead{}
\cfoot{\textit{AI Magazine}, 43(4), 344--352, 2022. \url{https://doi.org/10.1002/aaai.12068}
}

Text generation is a fundamental and long-lasting problem in the fields of natural language processing (NLP) and artificial intelligence~(AI), with wide applications such as conversational agents, news headline generation, and grammatical error correction. It is related to the foundation of AI, and is a key component in the Turing test~\cite{turing}. Early text generation systems are mainly based on rules and templates; thus, the generated text lacks flexibility and the applications are restricted to certain narrow domains, for example, report generation~\cite{report} and weather forecasting~\cite{weather,weather2}.

In recent years, deep learning has achieved remarkable progress in various text generation tasks~\cite{seq2seq,transformer}. Due to the high modeling capacity, deep neural networks are able to capture the complexity of language, and generate more diverse and natural texts than early rule-based systems. However, neural models are known to be data-hungry, usually requiring massive labeled input--output pairs, known as \textit{parallel corpora}. For example, the widely used corpus provided by the 2014 Workshop of Machine Translation contains more than 4 million pairs of English--German sentences~\cite{wmt14}. This is prohibitively expensive for deep learning models being applied to new domains, new tasks, and low-resource languages.

In this paper, I will introduce our recent progress on unsupervised text generation, which does not require parallel data for training. We formulate the text generation task as a search problem, where we define a heuristic objective function, typically involving language fluency, semantic coherency, and other task-specific constraints. Then, we perform discrete local search, for example, simulated annealing~\cite{upsa}, to generate the output sentence by maximizing the search objective. Further, a machine learning model may learn from the search results to smooth out the search noise and improve inference efficiency. In this way, our approach is unsupervised, yet achieving high performance in a variety of tasks. Empirically, our approach outperforms previous unsupervised text generation methods to a large extent, and has a comparable performance to supervised methods for text simplification~\cite{simplification}. Our study largely bridges the gap between supervised and unsupervised text generation. 

The rest of this paper is organized as follows: I will first discuss the social impacts of our work, followed by detailed but easy-to-understand technical explanation with examples and results. Then, I will mention a few important future directions that extend our current work, and finally conclude this paper.

\section{Social good}

Unsupervised text generation can benefit our society in various aspects.

First, unsupervised approaches are important to the AI industry, which typically requires that a minimal viable product is seen before investing substantial resources, including funding and human labor. Therefore, our unsupervised text generation techniques are particularly suitable for new tasks and startup companies, where massive annotated data are either unavailable or not affordable. 

Second, unsupervised text generation will help low-resource language processing. Training supervised deep learning NLP models requires massive labeled corpora, which only exist for several most-spoken languages. This prohibits the applications of the deep learning technique to low-resource languages. Our approaches, on the contrary, do not require parallel data, and can be potentially applicable to various text generation tasks in different languages.

Last but not least, our unsupervised techniques save human labor for data annotation. In machine learning, massive labeled data are often annotated by mechanical Turks, and this could be expensive and time-consuming; asking annotators to write sentences for text generation tasks is especially cumbersome and often yields poor-quality corpora. Our unsupervised text generation largely reduces human labor, as we do not require parallel data. Even if human-written text is needed for certain tasks, our approach is able to provide an initial draft for mechanical Turks to edit.

\section{Methodology}

We tackle unsupervised text generation by search approaches, where we heuristically define a scoring function that roughly estimates the quality of a candidate output sentence given some input in a certain task. Then, we perform discrete local search  to maximize the score for unsupervised text generation. This is accomplished by iteratively proposing a local edit of the candidate sentence, such as word insertion, deletion, and replacement. The proposed candidate may either be accepted or rejected depending on its score, although a better sentence is more likely to be accepted. In this way, we can gradually search for a high-scored sentence as the output text.

\subsection{Search objectives}

In a standard supervised machine learning application, the task is defined by data sets. For example, if the training corpus contains pairs of sentences of the same meaning but different wordings, then the trained machine learning model will accomplish the paraphrase generation task. However, defining a task by data sets is not feasible in the unsupervised setting, as we do not have parallel corpora. We observe that, in many text generation tasks, the quality of an output text can often be decomposed into several aspects, each of which can be modeled relatively easily. Thus, we may define the task by a heuristically designed scoring function, as explained below.

We would like the generated text to be fluent, so we use a language model to evaluate the language fluency of a given candidate sentence $\mathbf y$, denoted by $s_\text{fluency}(\mathbf y)$. A language model is trained to maximize the likelihood of a training corpus~\cite{slp}; the underlying assumption here is that if a sentence is more fluent, then its likelihood is higher, and vice versa. The language model can be  trained either on a task-specific unlabeled corpus using a recurrent neural network~\cite{rnnlm} or on large-scale generic corpora with the Transformer architecture~\cite{gpt}. Usually, fine-tuning a pre-trained language model on a task-specific corpus will yield the highest performance. It is noted that none of the above variants require labeled parallel corpora. 

Then, we consider the semantic of the generated sentence given some input. In various text generation tasks, the semantic of the output should be close to the input; examples of such applications include paraphrase generation and summarization. We leverage the \textit{embedding} technique~\cite{embedding}, which essentially maps an object (e.g., a word, a phrase, or a sentence) into a vector space. The cosine of two embedding vectors indicates how close two objects are. Again, the embeddings can be given by either recurrent neural networks~\cite{cosine} or by pre-trained language models~\cite{roberta}. Moreover, we find that in certain applications we may enhance the semantic measure by the embeddings of keywords~\cite{TGLS} and entities~\cite{simplification} to better preserve the content. In general, we denote the semantic scorer by $s_\text{semantic}(\mathbf y, \mathbf x)$ for a candidate output $\mathbf y$ given input $\mathbf x$.

There may be additional task-specific constraints, and we denote our third scorer by $s_\text{task}(\mathbf y)$. In fact, the task-specific constraint can be either a soft or a hard constraint, varying largely based on the nature of the task at hand. For keywords-to-sentence generation, $s_\text{task}(\mathbf y)$ is simply an indicator function, representing whether $\mathbf y$ contains the keywords or not~\cite{cgmh}; for text summarization, $s_\text{task}(\mathbf y)$ indicates whether the output is within the length budget or not~\cite{summarization}; and for paraphrasing, $s_\text{task}(\mathbf y)$ is counting the fraction of the overlap between input and output sentences~\cite{bleu}, as a paraphrase cannot be the same as the input.

We may also utilize external engineering for the task-specific constraint. The text simplification task appears similar to summarization, but emphasizes the simplicity of the output, instead of the length. In our study~\cite{simplification}, we adopt the Flesch reading ease score~\cite{ease}, which involves manually designed features, such as the sentence length and the number of syllables per word. Another interesting task is style-transfer generation, which aims to change the style of a sentence, while keeping the content, for example, changing an informal sentence to a formal one~\cite{formality}. In \newcite{TGLS}, we train a classifier in a supervised way based on style labels, serving as the task-specific scorer $s_\text{task}(\mathbf y)$.
It is noted that the style label is much easier to obtain than parallel corpora of sentence pairs, and our generation still works in an unsupervised manner.

In summary, our overall scoring function consists of language fluency, semantic similarity, and other task-specific constraints, having the following form
$$s(\mathbf y,\mathbf x)=s_\text{fluency}(\mathbf y)\cdot s_\text{semantic}(\mathbf y, \mathbf x)\cdot s_\text{task}(\mathbf y)$$
Here, we consider the multiplications of all the individual scorers because their scales may be different. If a scorer can be negative, it should be normalized to guarantee positivity. Generally speaking, our scorer estimates the fitness of a candidate sentence given an input for some task. It serves as our search objective for unsupervised text generation.

\subsection{Search algorithms}

We search towards the optimum of the objective function to obtain an output sentence. In supervised text generation, it is typical to apply beam search~\cite{seq2seq}, which searches for a word at a time but keeps a fix-sized beam to prevent the search space growing exponentially. This works well with the supervised machine learning model, where the joint probability of a sentence is decomposed into word-level probabilities, thus providing step-by-step guidance for the beam search. By contrast, our objective function yields a score only based on a complete sentence, instead of a partial sentence. Thus, traditional beam search is inappropriate in our scenario.

We observe, on the other hand, that the output and input resemble each other in various text generation applications. For example, a paraphrase is close to the original sentence, except that a few words and phrases are changed to synonyms and that the syntactic structure may be modified slightly; a summary even overlaps with the input text largely, although clutters and unimportant information are dropped. Thus, we propose to tackle unsupervised text generation by local search. In other words, we start from the input sentence and iteratively perform local editing to the candidate sentence to maximize the objective scoring function.

\begin{figure}[!t]
\centering
\includegraphics[width=.8\textwidth]{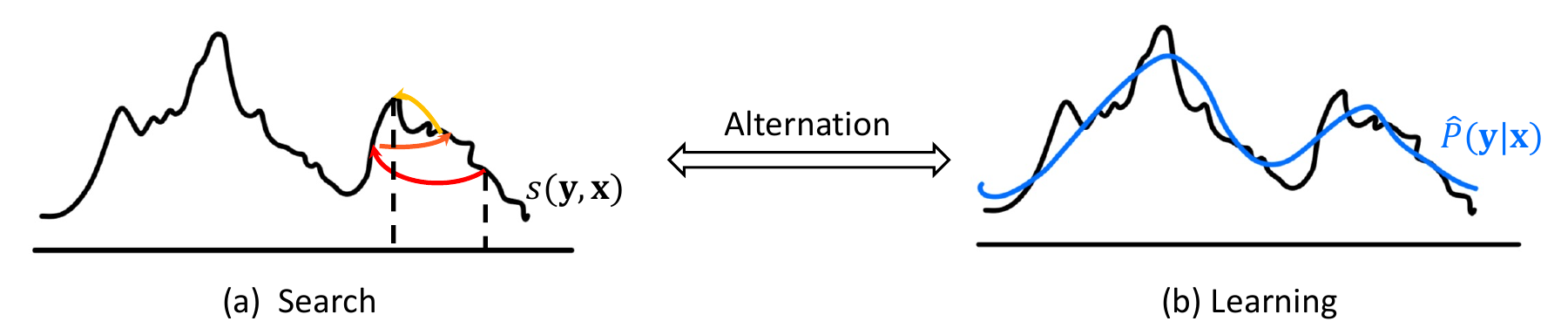}
\caption{Overview of our search-and-learning approach to unsupervised text generation. Alternations may be performed between search and learning to boost performance. The diagram is adapted from~\newcite{TGLS}.}
\label{fig}
\end{figure}

\begin{table}[!t]
    \centering
    \resizebox{\textwidth}{!}{
    \begin{tabular}{lllll}
    \toprule
         Task & Task-specific constraint & Edit operation & Algorithm & Study\\
    \midrule
         Grammatical error correction & None & Word editing & Metropolis--Hastings sampling & \newcite{cgmh}\\
         Paraphrasing & Different wordings & Word editing & Simulated annealing & \newcite{upsa}\\
         Summarization & Length & Word selection & Hill climbing & \newcite{summarization}\\
        Simplification & Ease score~\cite{ease} & Word/syntactical editing & Hill climbing &  \newcite{simplification}\\
         Style-transfer generation & Style accuracy & Word editing & Simulated annealing & \newcite{TGLS}\\
    \bottomrule
    \end{tabular}
    }
    \caption{Applications of unsupervised text generation.}
    \label{tab}
\end{table}

A local search algorithm is a two-stage process of proposal and acceptance. First, an edit operation is proposed from a candidate operation set, and then the algorithm will either accept or reject the proposal depending on the score of the resulting sentence. 

Our general operations of local editing include word replacement, insertion, and deletion. At each search step, we randomly select one of the edit operations. For deletion, we simply remove the selected word. For insertion and replacement, we need to propose a candidate word from the vocabulary. We propose to sample a word according to the Gibbs distribution~\cite{pgm} that is induced from our scorer, that is to say, a word that yields a better sentence should be more likely to be sampled. Such a proposal is more efficient than uniformly proposing a word from the vocabulary.

The editing operations may also be designed specifically to the task. In our study on text simplification~\cite{simplification}, we identify two major types of editing for simplifying a sentence: substituting a common word and re-ordering phrases. We use the WordNet~\cite{wordnet}, a lexical database, to obtain the synonyms, and substitute a rare word with a more frequently appearing synonym. For phrase re-ordering, we parse the sentence to a  constituency parse tree~\cite{corenlp}, which is a hierarchical organization of sentence components (e.g., noun phrases, verb phrases, and clauses). We manipulate the nodes of the parse tree for phrase re-ordering. It is noted that the proposed edit need not be perfect, as a corrupted sentence will be filtered out in the acceptance phrase.

Another simple yet interesting example of edit operation is for text summarization~\cite{summarization}. We notice that, in our application, a summary is oftentimes a subset of the input, even with the word order preserved. Therefore, we generate a summary by word-level extraction.
Suppose we have a budget of $k$ words for the summary, our system determines which $k$ words to be extracted from the input as the summary. This largely reduces the output space and simplifies the search process. Our edit operation is simply swapping the selection and non-selection of two words.

Regarding the acceptance stage, our first study~\cite{cgmh} adopts the Metropolis--Hastings algorithm~\cite{metropolis,mh}. This is essentially a sampling procedure that yields unbiased samples from the score-induced distribution. Although it is tempting to sample a sentence from a properly defined distribution for applications like paraphrasing, we later realize that optimization (rather than sampling) will lead to higher empirical performance due to better convergence; thus, we adopt search algorithms in our later studies.

Our mostly commonly used search algorithm is simulated annealing~\cite{sa}, which also works in a propose-and-accept manner~\cite{upsa,TGLS,dialog}. Generally, if the proposed sentence has a higher score, then the sentence is directly accepted. If the proposed sentence has a lower score, then it tends to be rejected. However, a worse sentence may still be accepted to better explore the search space, controlled by a temperature hyperparameter. If the temperature is high, then the algorithm is more exploratory, accepting more low-scored sentences. If the temperature is low, the algorithm is more greedy. Simulated annealing starts from a high temperature but gradually cools down during the search process, analogous to chemical annealing. This differs from Metropolis--Hastings sampling, because simulated annealing aims to converge to an optimum. 

In certain applications, a hill-climbing algorithm may also work well. Hill climbing is greedy search that only accepts better sentences. We adopt hill climbing for summarization, where we have design a simpler search space~\cite{summarization}, and for simplification, where we have dedicated engineering on edit operations~\cite{simplification}. 

Figure~\ref{fig}a illustrates the scoring function $s(\mathbf y, \mathbf x)$ and our search algorithm.  Table~\ref{tab} further summarizes the design of search objectives, edit operations, and search algorithms for different tasks in our studies.

\subsection{Learning from search results}
Admittedly, search-based text generation has its own disadvantages. First, our search objective is heuristically defined by several decomposed evaluation criteria. Although it correlates with output quality in the population level, the objective function may not be suitable for every single data sample.
Moreover, our algorithm is slow when deployed, because the search process requires several hundred iterations of proposals and re-evaluations. This prevents our approach from real-time applications.

To address the above drawbacks, we propose in \newcite{TGLS} to train a machine learning model that learns from the search results. Specifically, we adopt a pre-trained language model~\cite{gpt} and fine-tune it based on the input and the search algorithm's output to estimate the probability $\hat{P}(\mathbf{y}|\mathbf{x})$.

As known, a machine learning model pools together the knowledge of the individual samples, and thus is able to smooth out the noise of our heuristically defined objective function, demonstrated in Figure~\ref{fig}b. Empirically, our trained machine learning model immediately achieves higher performance than the search algorithm. In addition, this gives us 5--10 times speed-up in the deployment, because the machine learning model predicts an output sentence in a word-by-word fashion.

\begin{table}[!t]
    \centering
    \resizebox{\textwidth}{!}{
    \begin{tabular}{rl}
    \toprule
        Task & Keywords-to-sentence generation\cite{cgmh}\\
        Input & lottery, scholarships \\
        Output & But the lottery has provided
scholarships.\\
    \midrule
        Task & Grammatical error correction  \cite{cgmh}\\
        Input & Even if we are failed, we have to try to get a new things.\\
        Reference & Even if we all failed, we have to try to get new things.\\
        Output & Even if we are failing, we have to try to get some new things. \\
            \midrule
        
        Task & Paraphrasing~\cite{upsa}\\
        Input & Where are best places for spring snowboarding in the US? \\
        Reference & Where is the best place to snowboard in the US?\\
        Output & Where can I find the best places in the US for snowboarding?\\
                    \midrule

        Task & Summarization~\cite{summarization}\\
        Input & A German registered container ship ran aground at the entrance to the French port of Le Havre early Tuesday, but authorities said there were no casualties.\\
        Reference & Container ship runs aground in French port\\
        Output & A container ship ran aground but there were no casualties\\
    \bottomrule
    \end{tabular}}
    \caption{Examples of our unsupervised text generation in selected applications. References are the provided output sentences for the evaluation purpose; they are not used for training in our unsupervised setting.}
    \label{tab:examples}
\end{table}
\begin{figure}[!t]
    \centering
    \includegraphics[width=.45\textwidth]{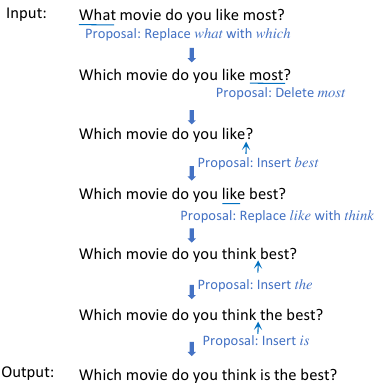}
    \caption{An example of the editing process for paraphrase generation~\cite{cgmh}. The rejected edits are omitted for clarity.}
    \label{fig:dynamics}
\end{figure}
Since our machine learning model outputs a better sentence than search, we may further feed it as the initial search candidate. It serves as a more meaningful starting point for the search algorithm than copying the input sentence. We alternate the search and learning processes to boost the performance.

Our idea of search and learning is similar to reinforcement learning~\cite{rl}. For example, the AlphaGo system performs Monte Carlo tree search for the Go game, and trains a neural network model from search results, outperforming world-class professional human Go players~\cite{mcts2,mcts}. We instead perform local search, such as simulated annealing, based on the characteristics of our own problems.

\subsection{Applications and results}

Our search-and-learning approach is a generic framework for unsupervised text generation, and can be applied to a variety of applications. Table~\ref{tab:examples} shows sample output in four tasks, namely, keywords-to-sentence generation, grammatical error correction, paraphrase generation, and summarization.
 For example, the sentence ``\textit{Where are best places for spring snowboarding in the US?}'' is paraphrased as ``\textit{Where can I find the best places in the US for snowboarding?}'' where our approach performs non-trivial editing. As seen, our approaches indeed generate meaningful output in each task, although it may be different from the reference output.

Figure~\ref{fig:dynamics} further demonstrates the dynamics of our editing in the paraphrase generation task. Given an input sentence ``\textit{What movie do you like most?}'' our approach suggests meaningful edits for paraphrasing, such as replacing \textit{what} with \textit{which} and replacing \textit{most} with \textit{best}. Although the intermediate sentences may not be perfect (e.g., ``\textit{Which movie do you think best?}''), our approach is able to further insert the words \textit{is the} to make the sentence fluent. The best candidate along the search process will be selected as the output.

\begin{table}[!t]
    \centering
    \resizebox{.7\textwidth}{!}{
    \begin{tabular}{llr}
    \toprule
        Category &  Method & iBLEU \\
    \midrule
        Supervised methods & \newcite{rlnn} & 14.83\\
        & Fine-tuning GPT-2~\cite{gpt} & 19.19\\
    \midrule
        Distant supervision & Round-trip translation & 14.36\\
                            & Domain adaptation~\cite{zichao}& 10.36\\
    \midrule
    Unsupervised methods & Variational autoencoder~\cite{vae} & 8.16\\
    & Our search approach & 14.52 \\
    & Our search and learning & 17.48\\
    \bottomrule
    \end{tabular}}
    \caption{Performance of paraphrase generation on a Quora dataset. The iBLEU metric measures the similarity against the reference, penalized by the similarity against the input; a higher score indicates a better model. Our search and learning results are quoted from \newcite{TGLS}. }
    \label{tab:results}
\end{table}

Quantitatively speaking, our search and learning approaches achieve high performance in all the above applications. Table~\ref{tab:results} lists the performance of our approaches compared with various competing models in different categories. As seen, both search-based and search-and-learning systems largely outperform the previous unsupervised paraphrase generator, given by a variational autoencoder~\cite{vae}. Although our unsupervised performance is worse than fine-tuning the pre-trained language model~\cite{gpt},  we outperform a recent competitive supervised system based on reinforcement learning~\cite{rlnn}. In text simplification, we see a similar trend that our unsupervised approach achieves comparable performance to supervised systems~\cite{simplification}.

In the industry, round-trip translation is widely used for paraphrasing, and in turn, for data augmentation. The idea is to make use of a machine translation system, translating a sentence into a foreign language and translating it back to obtain a paraphrase. This requires parallel data of translation rather than paraphrasing, known as \textit{distant supervision}. Our search-and-learning approach also largely outperforms the paraphrasing system based on round-trip translation. I fully believe that our search-and-learning framework will become a widely adopted industrial practice in the near future, as it largely bridges the gap between supervised and unsupervised text generation.

\section{Future directions}

The work described in this paper points to several important future directions of both fundamental and applied research.

One fundamental research question is how to perform efficient search in the sentence space, which can be inspired by the reinforcement learning community.
Our framework resembles the search and learning in reinforcement learning~\cite{rl}. However, text generation has its own challenges compared with other domains, such as videos games and the Go game. First, we have a larger branching factor than even the very challenging Go game; that is to say, for each generation step, we may have tens of thousands of words to consider from the entire vocabulary. Second, reinforcement learning highly depends on the reward function (i.e., the scoring function in our paper), which serves as the entire training signal. For example, the reward in a game could be the binary indicator of success or failure; in a typical reinforcement learning application, the reward function is often accurate, albeit noisy. However, our scoring function may be deterministically incorrect for certain samples,  as it is defined in a heuristic way. It remains unclear how these factors will affect the search process. 

\newcite{sha} proposes a gradient-based approach for searching in the lexical space. The gradient provides informative guidance for the proposal. However, Sha's current approach requires a differentiable scoring function, and it only works in the word level. It would be a promising direction to generalize the gradient-based editing in future work.

Another important extension is to propose edit operations beyond the word level. As seen in Figure~\ref{fig:dynamics}, local editing yields low-quality intermediate candidate sentences because our search algorithm may not be greedy, and this helps to explore a wide range of the sentence space. However, such editing proposals are more likely to be rejected due to the calculation of the acceptance rate. Therefore, it is tempting to propose editing in the phrase level, for example, replacing the phrase \textit{like most} to \textit{think is the best} in one step of editing. Although we have attempted syntax-based phrasal editing in \newcite{simplification}, our current treatment is still rudimentary as we only support phrase re-ordering. Future research may propose phrasal insertion and replacement by generating a plausible candidate phrase given a certain position in a sentence.

Currently, our approach is also restricted in that most of our tasks require the input and output closely resembling each other. This may be relaxed with additional heuristics based on the task. In unsupervised machine translation, for example, we may obtain an initial candidate translation by dictionary lookup of every word. Then, our approach can be applied to edit the candidate translation. 

For certain tasks, a completely unsupervised approach may not be feasible, because it is hard to decompose a scoring function into several aspects by heuristics. For example, it is difficult to articulate what constitutes a high-quality response in a dialogue system. Nevertheless, we may apply our search and learning approach to address other research questions. In a recent study of ours~\cite{dialog}, we adopt the search algorithm to control the emotion of a response. Traditional methods feed the desired emotion as input information, which may be ignored by the generation system~\cite{ecm}. By contrast, we directly edit a candidate response to ensure the presence of the desired emotion, thus generating much richer and more emotional responses than other dialogue systems. In \newcite{data2text}, we address few-shot data-to-text generation, where the goal is to generate a description based on tabular data with a few samples. We observe that, in the few-shot setting, important information is often missing in the output sentence. Thus, we adopt the search-and-learning approach, where we insert the missing information to improve semantic coverage, and further learn from the search results to improve fluency and efficiency. In this way, we only use 1\% of parallel data but achieve comparable performance to supervised learning with the entire data set.

Our work also opens wide applications beyond the natural language domain. In a recent study~\cite{SAGS}, we extend the simulated annealing to graph processing in the domain of molecule optimization. Our goal is to improve the hydrophobicity of a molecule, and we manage to design a scoring function that estimates hydrophobicity and molecule similarity; we further propose edit operations on graphs including node and edge insertion, deletion, and replacement. Experimental results show that, compared with previous molecule optimization methods, our approach significantly improves the hydrophobicity of a molecule by a series of local edits, while preserving other chemical characteristics.
This further shows the generality of our approach for different types of data and in different domains. The key to the applications is the design of search objective and edit operations. 

\section{Concluding remarks}

In this paper, I present our recent progress on search and learning approaches to unsupervised text generation, where we do not have parallel data for training. We define a heuristic scoring function that estimates the quality of the output text given input for a certain task. Then, we formulate text generation as a search problem, and adopt discrete search to maximize the objective score. We may further train a machine learning model from the search results to smooth out search noise and improve inference efficiency, while keeping the unsupervised nature. 

We have applied our framework to a variety of applications, including paraphrase generation, summarization, text simplification, keywords-to-sentence generation, grammatical error correction, and style-transfer generation. Our approach can be further adapted to other controllable text generation in the supervised or few-shot settings.  I also point out a few important future directions, including 
efficient search in the sentence space and applications to other domains beyond natural language.

Our search and learning approaches to unsupervised text generation will largely benefit the AI industry, because the unsupervised nature is helpful in building new products, especially for small businesses. Our work will also have positive social impacts in processing low-resource languages and saving human labor for data annotation.

\section{Acknowledgments}
The work was partially supported by the Natural Sciences and Engineering Research Council of Canada (NSERC) under grant No.~RGPIN-2020-04465, the Alberta Machine Intelligence Institute (Amii) Fellow Program, the Canada CIFAR AI Chair Program, Compute Canada (www.computecanada.ca), a UAHJIC project, and a donation from DeepMind.

\section{Biographical statement}

\textbf{Dr.~Lili Mou} is an assistant professor at the Department of Computing Science, University of Alberta. He is also an Alberta Machine Intelligence Institute (Amii) Fellow and a Canada CIFAR AI (CCAI) Chair. Lili received his BS and PhD degrees in 2012 and 2017, respectively, from School of EECS, Peking University. After that, he worked as a postdoctoral fellow at the University of Waterloo. His research interest is mainly in machine learning methods for natural language processing. He has more than 50 publications at top conferences and journals, such as AAAI, ACL, EMNLP, ICML, ICLR,  NeurIPS, and TACL. He has also presented conference tutorials at EMNLP-IJCNLP'19 and ACL'20. He received a New Faculty Highlight Award in the AAAI'21 conference.

\section{Conflict of interest}
The author has no conflicts of interest to report.

\bibliography{aaai}

\begin{thebibliography}{}

\bibitem[\protect\citeauthoryear{Bojar \bgroup et al\mbox.\egroup
  }{2014}]{wmt14}
Bojar, O.; Buck, C.; Federmann, C.; Haddow, B.; Koehn, P.; Leveling, J.; Monz,
  C.; Pecina, P.; Post, M.; Saint-Amand, H.; Soricut, R.; Specia, L.; and
  Tamchyna, A.
\newblock 2014.
\newblock Findings of the 2014 workshop on statistical machine translation.
\newblock In {\em Proceedings of the Ninth Workshop on Statistical Machine
  Translation},  12--58.

\bibitem[\protect\citeauthoryear{Bowman \bgroup et al\mbox.\egroup
  }{2016}]{vae}
Bowman, S.~R.; Vilnis, L.; Vinyals, O.; Dai, A.; Jozefowicz, R.; and Bengio, S.
\newblock 2016.
\newblock Generating sentences from a continuous space.
\newblock In {\em Proceedings of The 20th {SIGNLL} Conference on Computational
  Natural Language Learning},  10--21.

\bibitem[\protect\citeauthoryear{Dong \bgroup et al\mbox.\egroup
  }{2021}]{dialog}
Dong, C.; Huang, C.; Za{\"\i}ane, O.; and Mou, L.
\newblock 2021.
\newblock Simulated annealing for emotional dialogue systems.
\newblock In {\em Proceedings of the 30th ACM International Conference on
  Information and Knowledge Management},  2984--2988.

\bibitem[\protect\citeauthoryear{Goldberg, Driedger, and
  Kittredge}{1994}]{weather}
Goldberg, E.; Driedger, N.; and Kittredge, R.
\newblock 1994.
\newblock Using natural-language processing to produce weather forecasts.
\newblock {\em IEEE Expert} 9(2):45--53.

\bibitem[\protect\citeauthoryear{Hastings}{1970}]{mh}
Hastings, W.~K.
\newblock 1970.
\newblock {Monte Carlo} sampling methods using {Markov} chains and their
  applications.
\newblock {\em Biometrika} 57:970--109.

\bibitem[\protect\citeauthoryear{Jolly, Dengel, and Mou}{2021}]{data2text}
Jolly, S.; Dengel, A.; and Mou, L.
\newblock 2021.
\newblock Search and learn: Improving semantic coverage for data-to-text
  generation.
\newblock In {\em Proceedings of the Thirty-Sixth AAAI Conference on Artificial
  Intelligence},  10858--10866.

\bibitem[\protect\citeauthoryear{Jurafsky and Martin}{2009}]{slp}
Jurafsky, D., and Martin, J.~H.
\newblock 2009.
\newblock {\em Speech and Language Processing (2nd Edition)}.
\newblock Pearson.

\bibitem[\protect\citeauthoryear{Kincaid \bgroup et al\mbox.\egroup
  }{1975}]{ease}
Kincaid, J.~P.; Fishburne~Jr, R.~P.; Rogers, R.~L.; and Chissom, B.~S.
\newblock 1975.
\newblock Derivation of new readability formulas (automated readability index,
  fog count and flesch reading ease formula) for navy enlisted personnel.
\newblock Technical report, Naval Technical Training Command Millington TN
  Research Branch.

\bibitem[\protect\citeauthoryear{Koller and Friedman}{2009}]{pgm}
Koller, D., and Friedman, N.
\newblock 2009.
\newblock {\em Probabilistic Graphical Models: Principles and Techniques}.
\newblock MIT Press.

\bibitem[\protect\citeauthoryear{Kukich}{1983}]{report}
Kukich, K.
\newblock 1983.
\newblock Design of a knowledge-based report generator.
\newblock In {\em Proceedings of the 21st Annual Meeting of the Association for
  Computational Linguistics},  145--150.

\bibitem[\protect\citeauthoryear{Kumar \bgroup et al\mbox.\egroup
  }{2020}]{simplification}
Kumar, D.; Mou, L.; Golab, L.; and Vechtomova, O.
\newblock 2020.
\newblock Iterative edit-based unsupervised sentence simplification.
\newblock In {\em Proceedings of the 58th Annual Meeting of the Association for
  Computational Linguistics},  7918--7928.

\bibitem[\protect\citeauthoryear{Li \bgroup et al\mbox.\egroup }{2019}]{zichao}
Li, Z.; Jiang, X.; Shang, L.; and Liu, Q.
\newblock 2019.
\newblock Decomposable neural paraphrase generation.
\newblock In {\em Proceedings of the 57th Annual Meeting of the Association for
  Computational Linguistics},  3403--3414.

\bibitem[\protect\citeauthoryear{Li \bgroup et al\mbox.\egroup }{2020}]{TGLS}
Li, J.; Li, Z.; Mou, L.; Jiang, X.; Lyu, M.; and King, I.
\newblock 2020.
\newblock Unsupervised text generation by learning from search.
\newblock In {\em Advances in Neural Information Processing Systems},
  10820--10831.

\bibitem[\protect\citeauthoryear{Liu \bgroup et al\mbox.\egroup
  }{2019}]{roberta}
Liu, Y.; Ott, M.; Goyal, N.; Du, J.; Joshi, M.; Chen, D.; Levy, O.; Lewis, M.;
  Zettlemoyer, L.; and Stoyanov, V.
\newblock 2019.
\newblock {RoBERTa}: A robustly optimized {BERT} pretraining approach.
\newblock {\em arXiv preprint arXiv:1907.11692}.

\bibitem[\protect\citeauthoryear{Liu \bgroup et al\mbox.\egroup }{2020}]{upsa}
Liu, X.; Mou, L.; Meng, F.; Zhou, H.; Zhou, J.; and Song, S.
\newblock 2020.
\newblock Unsupervised paraphrasing by simulated annealing.
\newblock In {\em Proceedings of the 58th Annual Meeting of the Association for
  Computational Linguistics},  302--312.

\bibitem[\protect\citeauthoryear{Liu \bgroup et al\mbox.\egroup }{2021}]{SAGS}
Liu, X.; Li, P.; Meng, F.; Zhou, H.; Zhong, H.; Zhou, J.; Mou, L.; and Song, S.
\newblock 2021.
\newblock Simulated annealing for optimization of graphs and sequences.
\newblock {\em Neurocomputing} 465:310--324.

\bibitem[\protect\citeauthoryear{Manning \bgroup et al\mbox.\egroup
  }{2014}]{corenlp}
Manning, C.~D.; Surdeanu, M.; Bauer, J.; Finkel, J.~R.; Bethard, S.; and
  McClosky, D.
\newblock 2014.
\newblock The {Stanford CoreNLP} natural language processing toolkit.
\newblock In {\em Proceedings of 52nd Annual Meeting of the Association for
  Computational Linguistics: System Demonstrations},  55--60.

\bibitem[\protect\citeauthoryear{Metropolis \bgroup et al\mbox.\egroup
  }{1953}]{metropolis}
Metropolis, N.; Rosenbluth, A.~W.; Rosenbluth, M.~N.; Teller, A.~H.; and
  Teller, E.
\newblock 1953.
\newblock Equation of state calculations by fast computing machines.
\newblock {\em The Journal of Chemical Physics} 21(6):1087--1092.

\bibitem[\protect\citeauthoryear{Miao \bgroup et al\mbox.\egroup }{2019}]{cgmh}
Miao, N.; Zhou, H.; Mou, L.; Yan, R.; and Li, L.
\newblock 2019.
\newblock {CGMH}: Constrained sentence generation by {Metropolis-Hastings}
  sampling.
\newblock In {\em Proceedings of the AAAI Conference on Artificial
  Intelligence},  6834--6842.

\bibitem[\protect\citeauthoryear{Mikolov \bgroup et al\mbox.\egroup
  }{2010}]{rnnlm}
Mikolov, T.; Karafi{\'a}t, M.; Burget, L.; Cernock{\`y}, J.; and Khudanpur, S.
\newblock 2010.
\newblock Recurrent neural network based language model.
\newblock In {\em Proceedings of Interspeech},  1045--1048.

\bibitem[\protect\citeauthoryear{Mikolov \bgroup et al\mbox.\egroup
  }{2013}]{embedding}
Mikolov, T.; Sutskever, I.; Chen, K.; Corrado, G.~S.; and Dean, J.
\newblock 2013.
\newblock Distributed representations of words and phrases and their
  compositionality.
\newblock In {\em Advances in Neural Information Processing Systems},
  3111--3119.

\bibitem[\protect\citeauthoryear{Miller}{1995}]{wordnet}
Miller, G.~A.
\newblock 1995.
\newblock {WordNet}: A lexical database for english.
\newblock {\em Communications of the ACM} 38(11):39--41.

\bibitem[\protect\citeauthoryear{Pagliardini, Gupta, and Jaggi}{2018}]{cosine}
Pagliardini, M.; Gupta, P.; and Jaggi, M.
\newblock 2018.
\newblock Unsupervised learning of sentence embeddings using compositional
  n-gram features.
\newblock In {\em Proceedings of the 2018 Conference of the North {A}merican
  Chapter of the Association for Computational Linguistics: Human Language
  Technologies},  528--540.

\bibitem[\protect\citeauthoryear{Papineni \bgroup et al\mbox.\egroup
  }{2002}]{bleu}
Papineni, K.; Roukos, S.; Ward, T.; and Zhu, W.-J.
\newblock 2002.
\newblock {Bleu}: A method for automatic evaluation of machine translation.
\newblock In {\em Proceedings of the 40th Annual Meeting of the Association for
  Computational Linguistics},  311--318.

\bibitem[\protect\citeauthoryear{Qian \bgroup et al\mbox.\egroup }{2019}]{rlnn}
Qian, L.; Qiu, L.; Zhang, W.; Jiang, X.; and Yu, Y.
\newblock 2019.
\newblock Exploring diverse expressions for paraphrase generation.
\newblock In {\em Proceedings of the 2019 Conference on Empirical Methods in
  Natural Language Processing and the 9th International Joint Conference on
  Natural Language Processing},  3173--3182.

\bibitem[\protect\citeauthoryear{Radford \bgroup et al\mbox.\egroup
  }{2019}]{gpt}
Radford, A.; Wu, J.; Child, R.; Luan, D.; Amodei, D.; Sutskever, I.; et~al.
\newblock 2019.
\newblock Language models are unsupervised multitask learners.
\newblock {\em OpenAI Blog}.

\bibitem[\protect\citeauthoryear{Rao and Tetreault}{2018}]{formality}
Rao, S., and Tetreault, J.
\newblock 2018.
\newblock Dear sir or madam, may {I} introduce the {GYAFC} dataset: Corpus,
  benchmarks and metrics for formality style transfer.
\newblock In {\em Proceedings of the 2018 Conference of the North {A}merican
  Chapter of the Association for Computational Linguistics: Human Language
  Technologies},  129--140.

\bibitem[\protect\citeauthoryear{Reiter \bgroup et al\mbox.\egroup
  }{2005}]{weather2}
Reiter, E.; Sripada, S.; Hunter, J.; Yu, J.; and Davy, I.
\newblock 2005.
\newblock Choosing words in computer-generated weather forecasts.
\newblock {\em Artificial Intelligence} 167:137--169.

\bibitem[\protect\citeauthoryear{Saygin, Cicekli, and Akman}{2000}]{turing}
Saygin, A.~P.; Cicekli, I.; and Akman, V.
\newblock 2000.
\newblock Turing test: 50 years later.
\newblock {\em Minds and Machines} 10(4):463--518.

\bibitem[\protect\citeauthoryear{Schumann \bgroup et al\mbox.\egroup
  }{2020}]{summarization}
Schumann, R.; Mou, L.; Lu, Y.; Vechtomova, O.; and Markert, K.
\newblock 2020.
\newblock Discrete optimization for unsupervised sentence summarization with
  word-level extraction.
\newblock In {\em Proceedings of the 58th Annual Meeting of the Association for
  Computational Linguistics},  5032--5042.

\bibitem[\protect\citeauthoryear{Sha}{2020}]{sha}
Sha, L.
\newblock 2020.
\newblock Gradient-guided unsupervised lexically constrained text generation.
\newblock In {\em Proceedings of the 2020 Conference on Empirical Methods in
  Natural Language Processing},  8692--8703.

\bibitem[\protect\citeauthoryear{Silver \bgroup et al\mbox.\egroup
  }{2017}]{mcts2}
Silver, D.; Schrittwieser, J.; Simonyan, K.; Antonoglou, I.; Huang, A.; Guez,
  A.; Hubert, T.; Baker, L.; Lai, M.; Bolton, A.; et~al.
\newblock 2017.
\newblock Mastering the game of {Go} without human knowledge.
\newblock {\em Nature} 550(7676):354--359.

\bibitem[\protect\citeauthoryear{Silver \bgroup et al\mbox.\egroup
  }{2018}]{mcts}
Silver, D.; Hubert, T.; Schrittwieser, J.; Antonoglou, I.; Lai, M.; Guez, A.;
  Lanctot, M.; Sifre, L.; Kumaran, D.; Graepel, T.; et~al.
\newblock 2018.
\newblock A general reinforcement learning algorithm that masters chess, shogi,
  and {Go} through self-play.
\newblock {\em Science} 362(6419):1140--1144.

\bibitem[\protect\citeauthoryear{Sutskever, Vinyals, and Le}{2014}]{seq2seq}
Sutskever, I.; Vinyals, O.; and Le, Q.~V.
\newblock 2014.
\newblock Sequence to sequence learning with neural networks.
\newblock In {\em Advances in Neural Information Processing Systems},
  3104--3112.

\bibitem[\protect\citeauthoryear{Sutton and Barto}{2018}]{rl}
Sutton, R.~S., and Barto, A.~G.
\newblock 2018.
\newblock {\em Reinforcement Learning: An Introduction}.
\newblock MIT Press.

\bibitem[\protect\citeauthoryear{Van~Laarhoven and Aarts}{1987}]{sa}
Van~Laarhoven, P.~J., and Aarts, E.~H.
\newblock 1987.
\newblock {\em Simulated Annealing: Theory and Applications}.
\newblock Springer.

\bibitem[\protect\citeauthoryear{Vaswani \bgroup et al\mbox.\egroup
  }{2017}]{transformer}
Vaswani, A.; Shazeer, N.; Parmar, N.; Uszkoreit, J.; Jones, L.; Gomez, A.~N.;
  Kaiser, {\L}.; and Polosukhin, I.
\newblock 2017.
\newblock Attention is all you need.
\newblock In {\em Advances in Neural Information Processing Systems},
  5998--6008.

\bibitem[\protect\citeauthoryear{Zhou \bgroup et al\mbox.\egroup }{2018}]{ecm}
Zhou, H.; Huang, M.; Zhang, T.; Zhu, X.; and Liu, B.
\newblock 2018.
\newblock Emotional chatting machine: Emotional conversation generation with
  internal and external memory.
\newblock In {\em Proceedings of the AAAI Conference on Artificial
  Intelligence},  730--738.

\end{thebibliography}

\end{document}